# Segmentation of Camera Captured Business Card Images for Mobile Devices

Ayatullah Faruk Mollah
School of Mobile Computing and Communication
Jadavpur University
Kolkata, India
afmollah@gmail.com

Subhadip Basu, Mita Nasipuri
Department of Computer Science & Engineering
Jadavpur University
Kolkata, India
{subhadip,mnasipuri}@cse.jdvu.ac.in

*Abstract*—Due to huge deformation in the camera captured images, variety in nature of the business cards and the computational constraints of the mobile devices, design of an efficient Business Card Reader (BCR) is challenging to the researchers. Extraction of text regions and segmenting them into characters is one of such challenges. In this paper, we have presented an efficient character segmentation technique for business card images captured by a cell-phone camera, designed in our present work towards developing an efficient BCR. At first, text regions are extracted from the card images and then the skewed ones are corrected using a computationally efficient skew correction technique. At last, these skew corrected text regions are segmented into lines and characters based on horizontal and vertical histogram. Experiments show that the present technique is efficient and applicable for mobile devices, and the mean segmentation accuracy of 97.48% is achieved with 3 mega-pixel (500-600 dpi) images. It takes only 1.1 seconds for segmentation including all the preprocessing steps on a moderately powerful notebook (DualCore T2370, 1.73 GHz, 1GB RAM, 1MB L2 Cache).

I. INTRODUCTION

Though mobile devices such as Personal Digital Assistants (PDA) and cell phones are very popular for the purpose of managing contact profiles, business cards have not lost its popularity. Rather, it is being used even more than ever before. Its usage is no more limited to business groups but is extensively used for non-commercial purposes also. Teachers, doctors, professors, activists and individuals use business cards (also known as name card or visiting card). It has been a tool for presenting one's profile, building personal and professional relationship, means of advertising, etc. But, as the number of such cards becomes very large, a person face a lot of difficulties in managing them, getting the right contact at the right time and at the right place. In order to ease this problem, business card album has been used so far to store and organize the business cards. The disadvantages of managing business cards with an album are that it takes considerable time to find a card from an album and more importantly one can not always carry the album with him/her. In addition to that, one has to type the required information from the card before solving the purpose. For instance, to make a phone call one has to retrieve and type the number in a cell phone before making the call.

On the other hand, most of the handheld mobile devices support full profiles of each person for the contact books. So, if the required information of the business cards can be populated into the contact book through an Optical Character Recognition (OCR) system, it would be very handy and useful for the purpose. Such an OCR system is commercially termed as Business Card Reader (BCR). A number of cell-phones such as 'Motorola MOTOROKR E6', 'Sony Ericsson P1i' and 'ASUS P527' have pre-installed BCR softwares. A downloadable BCR software developed by Abbyy can be downloaded from [1] and installed on some specified cell-phones. But the accuracy is not satisfactory so as to be really useful in practice.

Developing an efficient BCR involves a number of challenges mentioned below.

- Deformation of acquired images

  Business card images as acquired with the camera of a handheld device are often distorted and degraded. If the imaging device is not on the normal of the object plane, perspective distortion occurs due to the perspective projection of the object on the camera plane. When but both the camera plane and the object plane are parallel but the axis of the camera and that of the object are not parallel to each other, the acquired image becomes skewed. While capturing images of planar objects like business cards, we observe that skew and perspective distortions are very common due to manual estimation and adjustment. Therefore, one may need to correct the skew and perspective distortion before segmenting.

- Degradation of acquired images

  Similar to distortion, the camera captured images may be degraded due to lighting intensity of the environment, blur, imperfect focus, shadow, uneven illumination as mentioned in [2], use of the flash light while capturing the image, and gray scale conversion of the color cards. It is observed that if camera flash is used, the point of flash focus becomes brighter and the intensity decays outward [3].

- Multiplicity in nature of business cards

  Besides distortion and degradation, business card images are of multiple natures. There might be pictures and graphics including logo at either or both background and foreground. Varieties of fonts of different size are widely used. The closely placed words and lines of a text region may also get connected that makes the segmentation difficult.

- Computational Constraints of cell-phones

  While designing any algorithm for mobile devices, one must keep in mind that the cell-phones usually have low computing power (200-333 MHz ARM series processors), limited caching and small primary memory (upto 128 MB). Cell-phones do not have Floating Point Unit (FPU) which is used for floating point operations. So, the techniques/algorithms must be computationally efficient and light weight. The less is the number of floating point operations, the better is the technique. Computationally expensive algorithms, however well be their performances, can not be implemented on such low computing platform for practical solutions.

One of the most important steps of designing an efficient BCR system is to segment the business cards properly. Improper segmentation affects the recognition process and thus the overall performance of the system. So, the non-text elements such as logo, images, graphics, etc. must be eliminated at first. Then lines, words and characters are to be segmented from the text regions.

Literature reveals a number of character segmentation algorithms/techniques most of which are for document images. Keeping the computational limitation of the mobile devices in view, simple projection based approach has been implemented in [4-5]. In this approach, an assumption is made that only one text line is present along a horizontal direction. So, horizontal histogram profile is used to identify lines and vertical histogram profile is taken into account to segment characters from these lines. This approach is very common and easy to implement. The accuracy of this technique depends on the pre-processing i.e. text/background separation, binarization and skew correction.

Wang et al. [6] has presented a character segmentation method for color images. At first, the edges are detected with the help of color differences and then the watershed transform is applied to segment the regions. The segmented regions are then clustered based on their characteristics. Based on the clustering, characters are selected using connected component analysis. Log-Gabor filter based segmentation approaches are found in [7-8]. In these approaches, Log-Gabor filter is used to take the advantages of both the gray level variation and the special location. In [7], recognition is also incorporated to determine some of the parameters needed in the filter. Another segmentation technique has been presented by Shi et al. [9] using background light intensity normalization for handwritten historical document images. However, the above segmentation techniques may not always be fit for the low resource computing devices.

In this paper, we have presented and evaluated a technique for fast segmentation of the extracted text regions from business card images. At first, the text regions are extracted from the input images as discussed in Section II.A. Then the text regions are skew corrected as illustrated in Section II.B. Skew corrected text regions are then binarized and then segmented into lines and characters as discussed in Section II.C and Section II.D respectively.

II. THE PRESENT WORK

The business card image is directly passed to the engine for extracting text regions without any kind of enhancement or correction. Once the text regions are identified and extracted, the skew angle for them are determined and rectified. After that the text regions are segmented into lines and characters.

A. *Text Region Extraction*

A business card image is split into rectangular blocks that are classified as part of foreground or background based on the intensity variance within the block. After doing so, the foreground components including texts are isolated from the background. Then the non-text components among them such as logos, lines, graphics, pictures and noises are eliminated. After that, well spaced lines of the card form separate connected components (CC) or regions but most of the skewed and closely-placed neighboring lines together form a single CC. Fig. 1 show two sample business card images and the extracted text regions from them. The detail of the technique has been published in our previous work [10].

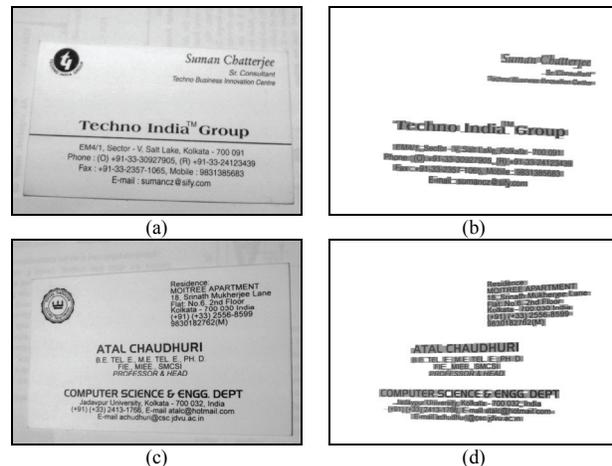

(a)      (b)

(c)      (d)

Figure 1. Text Region Extraction. (a) An unevenly skewed card image, (b) Extracted text regions for card (a), (c) Sample business card image having text regions skewed by both positive and negative angle, (d) Extracted text regions for card (c)

B. *Skew Angle Computation*

Skew angle is estimated for each text region and rotate it accordingly. To calculate the skew angle, we consider the bottom profile of the gray shade of a text region. The profile contains the height in terms of pixel from the bottom edge of the rectangle

International Journal of Computer Science and Applications, 1(1): 33-37

formed by the text region to the first gray/black pixel found while moving upward. A detail discussion of the technique can be found in [11]. However, a brief overview is given here.

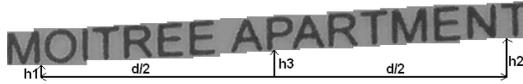

Figure 2. Skew Angle Computation

As the profile is ready, we calculate the mean ($\mu$) and the mean deviation ($\tau$) as shown in Eq. 1 and 2 respectively. The computation of mean deviation does not involve floating point arithmetic. Although, we can convert the floating point arithmetic to integer one, we want to avoid it as our intent is to embed the method on mobile devices having no FPUs. Some elements of the profile that are not in sync with the others i.e. not within (+$\tau$, -$\tau$) are removed.

$$\mu = \frac{1}{N}\sum_{i=0}^{N-1} h[i] \quad (1)$$

$$\tau = \frac{1}{N}\sum_{i=0}^{N-1} |\mu - h[i]| \quad (2)$$

where *N* is the profile length and *h* is the profile array

Among the remaining profile elements that really contribute to the actual skew of the text regions, we consider the leftmost (*h1*), rightmost (*h2*) and the middle profile element (*h3*) as shown in Fig. 2. The distance between *h1* and *h2* is computed as *d*. Then, the individual skew angles for the slope between *h1* and *h2*($\alpha$), *h1* and *h3*($\beta$), and *h2* and *h3*($\gamma$) are computed as formulated in Eq. 3-5 respectively. Now, ideally they should be the same. A threshold ($\varepsilon$) is introduced to allow a certain deviation in between them. So, if no deviation in between $\alpha$, $\beta$ and $\gamma$ is more than $\varepsilon$, we take an average and rectify the skew of the text region. Otherwise, we look forward to the top profile of the text region and compute the skew angle. Respective skew angles as computed from the top profile of the text region are $\alpha'$, $\beta'$ and $\gamma'$. If these are found to be inline, we take an average of them and rectify the skew. Else, the smaller one between the averages obtained from top and bottom profiles is considered as the skew angle. It may be noted that this approach gives a mean to bypass some computation if not required.

$$\alpha = \arctan\left(\frac{\delta h}{d}\right), \text{ where } \delta h = h2 - h1 \quad (3)$$

$$\beta = \arctan\left(\frac{\delta h}{d}\right), \text{ where } \delta h = h3 - h1 \quad (4)$$

$$\gamma = \arctan\left(\frac{\delta h}{d}\right), \text{ where } \delta h = h2 - h3 \quad (5)$$

C. *Binarization*

Document image binarization techniques can not be directly implemented on mobile devices because of their computational requirements. Even the global binarization technique such as Otsu[12] seems to be computationally expensive for low computing architectures. Conventional adaptive binarization techniques like [13-14] have no way to directly implement on mobile devices unless they are simplified significantly. It is well known that the adaptive binarization techniques usually yield better results than the global ones. In our present work, we have designed a computationally efficient yet adaptive binarization technique which is detailed in [15]. This technique is found to be satisfactory for the applications like BCR that are supposed to run on mobile devices.

D. *Character Segmentation*

A text region extracted with the present technique may have multiple lines. We segment the text regions into text lines and then characters are segmented from them. Horizontal histogram profile is analyzed for line segmentation. All possible line segments are determined by comparing the profile elements with a considerably large threshold. After that the inter-segment distances are analyzed and some segments are rejected. The idea behind this technique is that the distance in terms of pixel between two lines will not be too small and the inter-segment distances are likely to become equal.

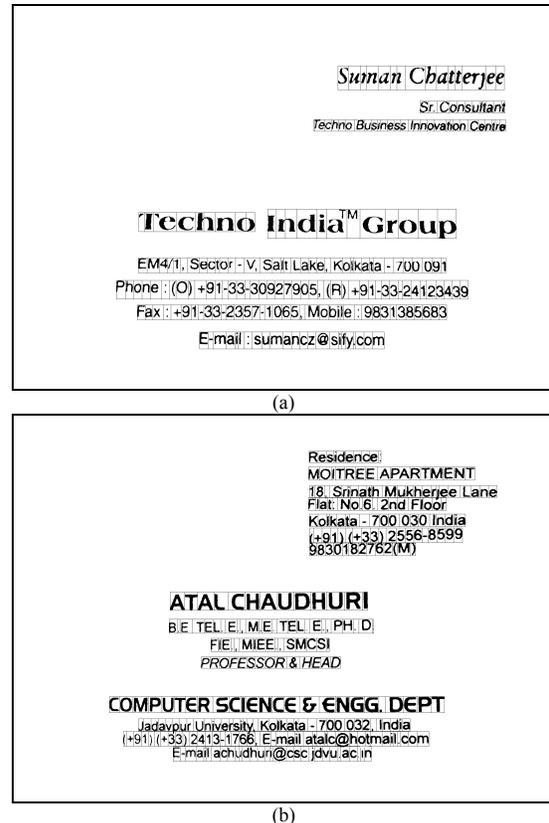

Figure 3. Character segmentation for some sample business card images. (a) Segmented characters for Fig. 1(a) are marked with boxes, (b) Segmented characters for Fig. 1(c) are also marked with boxes

Once a line is segmented, characters segmentation is done simply on the basis of vertical histogram profile.

### III. EXPERIMENTAL RESULTS

Experiments have been carried out to evaluate the performance and applicability of the current technique for character segmentation on a set of 100 business card images of wide variety. The images have been captured with a cell phone (Sony Ericsson K810i) camera.

Character segmentation accuracy has been estimated in terms of the ratio of the number of correct segmented characters to the total number of characters present in a card image. A character is categorized as incorrectly segmented one if it is over-segmented or is segmented as a part of another character.

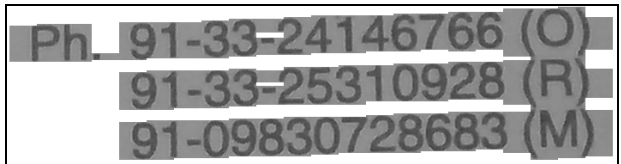
Figure 4. Sample text regions before segmentation

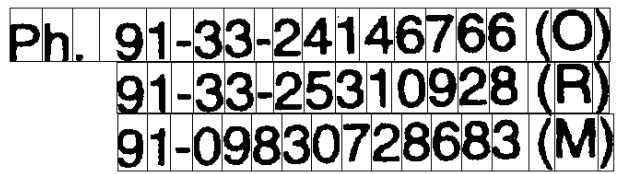
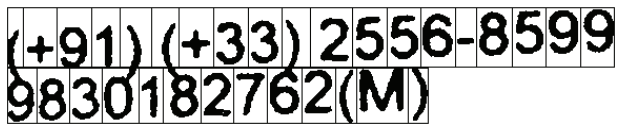
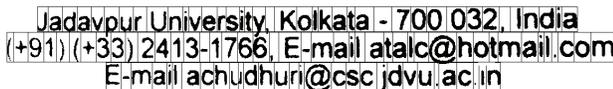
Figure 5. Segmented characters for the respective images of Fig. 4

By following this estimation technique we found that the character segmentation accuracy is 97.48% in case of 3 MP images. It may be noted that the present segmentation technique is not meant for italic and cursive texts. So, such texts have been ignored while calculating the segmentation accuracy.

Segmented characters from the Fig. 1(a) and (c) have been shown in Fig. 3(a) and (b) respectively. The rectangular blocks denote the segmented characters for each line. Fig. 4 shows some sample skewed text regions containing multiple lines. Segmented characters of these text regions have been shown in Fig. 5.

The applicability of the present technique on mobile devices is tested in terms of computational complexity. The average time consumption with respect to a moderately powerful computer (DualCore T2370, 1.73 GHz, 1GB RAM, 1MB L2 Cache) is 1.1 second for 3 MP images. It may be noted that the technique seems to be fast enough.

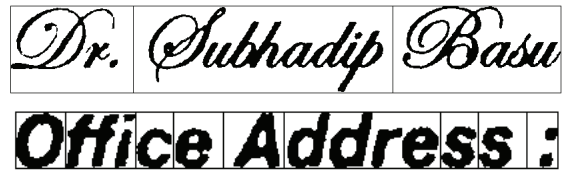
Figure 6. Segmentation failures

However, the technique has certain limitations too. It cannot work well with italic or cursive text lines or words as shown in Fig. 6.

### IV. CONCLUSION

One may think the present segmentation technique along with pre-processing is trivial. But, keeping in view the computational constraints of the mobile devices and the computational requirements of the existing techniques, we may say that the present technique is a step forward towards segmenting image embedded text documents like business card images. It seems to be efficient subject to applicability on low computing architectures. The present technique is also applicable for any image embedded text documents. However, there are limitations that reflect for cursive and italic texts. As our approach is top-down, we get a text region at first, then the text lines of the text region and at last the characters segmented from the text lines. Thus, the technique has a potential advantage for organizing the individual segmented characters and thus preserving the layout to a great extent. Our future work aims at recognizing the segmented characters and rearrange them in order to classify the contents such as 'name', 'telephone number', 'email', etc. Other than the BCR and general OCR applications, the present approach can also be deployed for textual Content Based Image Retrieval (CBIR) systems.

### ACKNOWLEDGEMENT

Authors are thankful to the *Center for Microprocessor Application for Training Education and Research (CMATER)* and project on *Storage Retrieval and Understanding of Video for Multimedia (SRUVM)* of the Department of Computer Science and Engineering, Jadavpur University for providing infrastructural support for the research work. We are also thankful to the *School of Mobile Computing and Communication (SMCC)* for proving the research fellowship to the first author.

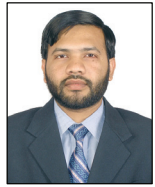

**Ayatullah Faruk Mollah**
Research Fellow
School of Mobile Computing and Communication
Jadavpur University
afmollah@gmail.com

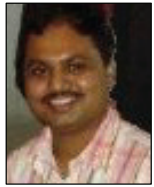

**Subhadip Basu**
Senior Lecturer
Department of Computer Science and Engineering
Jadavpur University
subhadip@cse.jdvu.ac.in

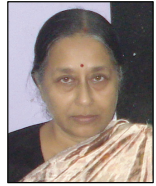

**Mita Nasipuri**
Professor
Department of Computer Science and Engineering
Jadavpur University
mnasipuri@cse.jdvu.ac.in